\documentclass[runningheads]{llncs}
\usepackage{graphicx}
\usepackage{tikz}
\usepackage{comment}
\usepackage{amsmath,amssymb} 
\usepackage{color}
\usepackage[accsupp]{axessibility}  

\usepackage{multirow}
\usepackage{graphicx,xparse,booktabs}
\usepackage{makecell}
\usepackage{pifont}
\newcolumntype{x}[1]{>{\centering\arraybackslash\hspace{0pt}}p{#1}}
\newcommand{\PreserveBackslash}[1]{\let\temp=\\#1\let\\=\temp}
\newcolumntype{C}[1]{>{\PreserveBackslash\centering}p{#1}}
\newcolumntype{L}[1]{>{\PreserveBackslash\raggedright}p{#1}}
\usepackage[pagebackref=true,breaklinks=true,colorlinks,bookmarks=true]{hyperref} 
\usepackage[misc]{ifsym}

\begin{document}
\pagestyle{headings}
\mainmatter
\def\ECCVSubNumber{1729}  

\title{Registration based Few-Shot Anomaly Detection} 

\titlerunning{Registration based Few-Shot Anomaly Detection}
\author{Chaoqin Huang\inst{1,3,4} \and
Haoyan Guan\inst{2} \and
Aofan Jiang\inst{1} \and
Ya Zhang\inst{1,3} \textsuperscript{\Letter} \and \\
Michael Spratling\inst{2} \and
Yan-Feng Wang\inst{1,3} \textsuperscript{\Letter}}
\authorrunning{C. Huang et al.}
\institute{Cooperative Medianet Innovation Center, Shanghai Jiao Tong University 
\email{\{huangchaoqin, stillunnamed, ya\_zhang, wangyanfeng\}@sjtu.edu.cn}
\and
King's College London \\
\email{\{haoyan.guan, michael.spratling\}@kcl.ac.uk}
\and
Shanghai Artificial Intelligence Laboratory
\and
National University of Singapore
}
\maketitle

\begin{abstract}
This paper considers few-shot anomaly detection (FSAD), a practical yet under-studied setting for anomaly detection (AD), where only a limited number of normal images are provided for each category at training. So far, existing FSAD studies follow the one-model-per-category learning paradigm used for standard AD, and the inter-category commonality has not been explored. Inspired by how humans detect anomalies, \emph{i.e.,} comparing an image in question to normal images, we here leverage registration, an image alignment task that is inherently generalizable across categories, as the proxy task, to train a category-agnostic anomaly detection model. During testing, the anomalies are identified by comparing the registered features of the test image and its corresponding support (normal) images. As far as we know, this is the first FSAD method that trains a single generalizable model and requires no re-training or parameter fine-tuning for new categories. Experimental results have shown that the proposed method outperforms the state-of-the-art FSAD methods by 3\%-8\% in AUC on the MVTec and MPDD benchmarks. Source code is available at: \url{https://github.com/MediaBrain-SJTU/RegAD}
\end{abstract}

\begin{figure}[t]
\centering
\includegraphics[width=0.95\textwidth]{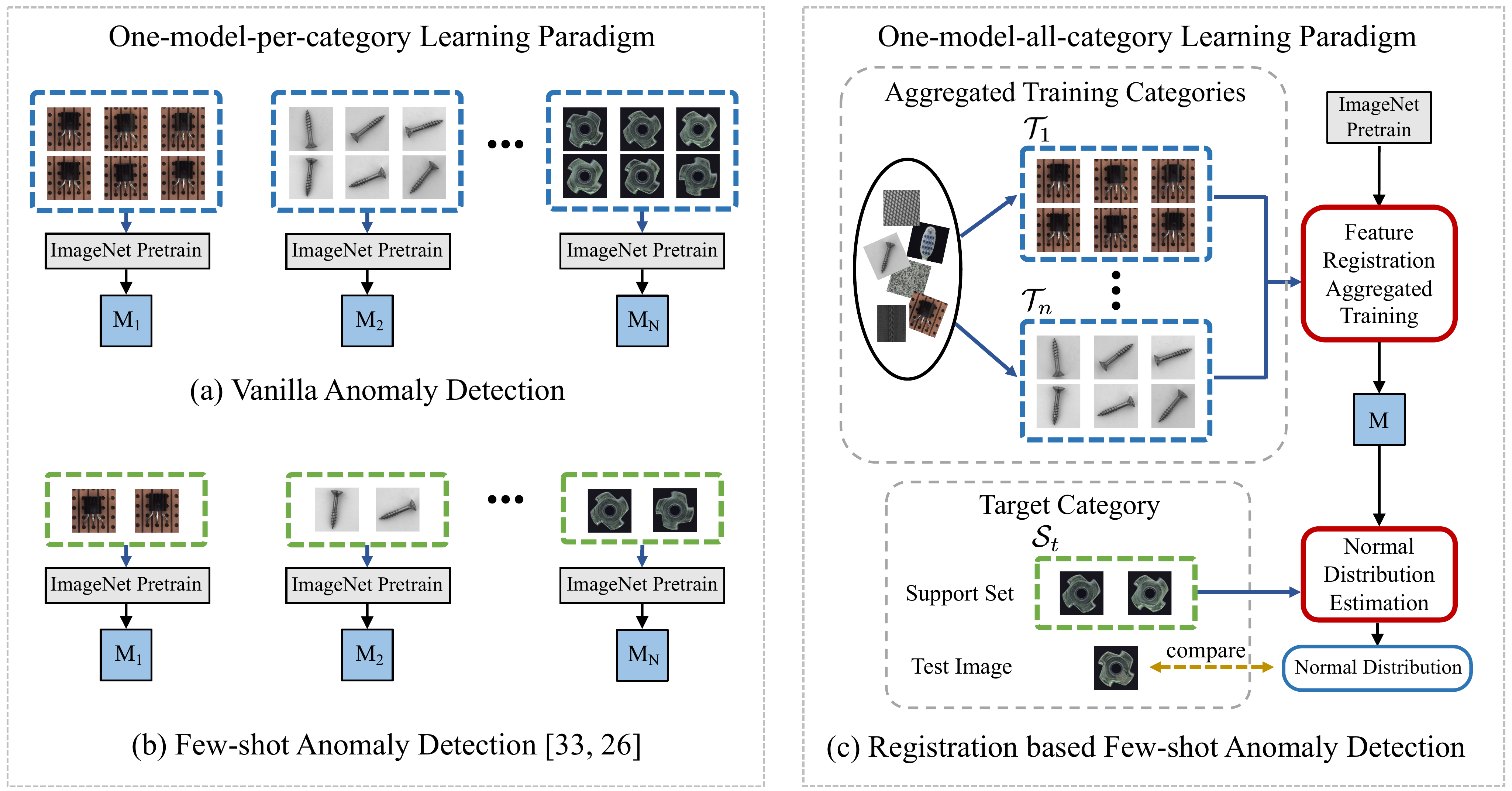}
\caption{Different from (a) vanilla AD, and (b) existing FSAD methods under the one-model-per-category learning paradigm, the proposed method (c) leverages feature registration as a category-agnostic approach for FSAD, under the one-model-all-category learning paradigm. Trained with aggregated data of multiple categories, the model is directly applicable to novel categories without any parameter fine-tuning, with the only need to estimate the normal feature distribution given the corresponding support set.}
\label{img:intro}
\end{figure}

\section{Introduction}
Anomaly detection (AD), with a wide range of applications such as defect detection~\cite{matsubara2018anomaly}, medical diagnosis~\cite{zhang2020viral}, and autonomous driving~\cite{eykholt2018robust}, has received quite some attention in the computer vision community over the last decades. With the ambiguous definition of ``anomaly'', \emph{i.e.,} samples that do not conform to the ``normal'', it is impossible to train with an exhaustive set of anomalous samples. As a result, recent studies on anomaly detection have largely been devoted to unsupervised learning, \emph{i.e.,} learning with only the ``normal'' samples. Through modeling the normal distribution with one-class classification~\cite{scholkopf2001estimating,ruff2018deep,yi2020patch}, reconstruction~\cite{zong2018deep,gong2019memorizing,metaformer,huang2022esad}, or self-supervised learning tasks~\cite{golan2018deep,ARNet,MKD,focus}, many AD methods detect anomalies by identifying samples with different distributions than the model.

Most existing AD methods have focused on training a dedicated model for each category (Fig.~\ref{img:intro} (a)). However, in real-world scenarios such as defect detection, given hundreds of industrial products to handle, it is not cost-effective to collect a large training set for each product, not to mention the need for many time-sensitive applications. A couple of studies~\cite{TDG,DiffNet} have recently explored a special, yet practical, setting of AD, \emph{i.e.,} few-shot anomaly detection (FSAD), where only a limited number of normal images are provided for each category at training (Fig.~\ref{img:intro} (b)). The few-shot learning of anomaly detection has been approached with strategies to reduce the demand on training samples, such as radical data augmentation with multiple transformations~\cite{TDG} or a lighter estimator for the normal distribution estimation~\cite{DiffNet}. \emph{However, such approaches still follow the one-model-per-category learning paradigm and fail to leverage the inter-category commonality.}

This paper aims to explore a new paradigm for FSAD, by learning a common model shared among multiple categories and also generalizable to novel categories, and inspired by how human beings detect anomalies. In fact, when a human is asked to search for the anomaly in an image, a simple strategy one may adopt is to compare the sample to a normal one to find the difference. As long as one knows how to compare two images, the actual semantics of the images does not matter anymore. To achieve such a human-like comparison process, we resort to registration, a process of transforming different images into one coordinate system in order to better enable comparison~\cite{brown1992survey,Barbara2003Image,peng2011brainaligner}. Registration is particularly suitable for FSAD, as \emph{registration is expected to be category-agnostic and thus generalizable across categories, allowing the model to be adaptable to novel categories without the necessity of parameter fine-tuning.}

Fig.~\ref{img:intro} (c) provides an overview of the proposed \underline{Reg}istration based few-shot \underline{A}nomaly \underline{D}etection (RegAD) framework. To train a category-agnostic anomaly detection model, we leverage registration, a task that is inherently generalizable across categories, as the proxy task. A Siamese network~\cite{chen2021exploring} with three spatial transformer network~\cite{STN} blocks is employed as the registration network (see Fig.~\ref{img:RegAD}). For better robustness, instead of registering the images pixel-by-pixel as typical registration methods~\cite{peng2011brainaligner}, here we propose a feature-level registration loss by maximizing the cosine similarity of features from the same category, which may be deemed as a relaxed version of the pixel-wise registration loss. Normal images from different categories are used together to aggregately train the model, with two images from the same category randomly selected as a training pair. Such aggregated training procedure is adopted so as to enable the trained registration model to be category-agnostic. At test time, a support set of a few normal samples is provided for the target category, together with each test sample. It is straightforward to identify anomalies by comparing the registered features of the test image and the corresponding support (normal) images. Given the support set, the normal distribution of registered features for the target category is estimated with a statistical-based distribution estimator~\cite{defard2021padim}. Test samples that are out of the statistical normal distribution are considered anomalies. In this way, the model quickly adapts to novel categories by simply estimating its normal feature distribution without any parameter fine-tuning.

To validate the effectiveness of RegAD, we experiment with two challenging benchmark datasets for industrial defect detection, MVTec AD~\cite{bergmann2019mvtec} and MPDD~\cite{jezek2021deep}. Our experimental results have shown that RegAD outperforms the state-of-the-art FSAD methods~\cite{TDG,DiffNet}, achieving improvements of 5.1\%, 6.9\%, and 8.0\% in AUC on MVTec, and improvements of 3.2\%, 5.0\%, and 3.4\% in AUC on MPDD, for 2-shot, 4-shot, and 8-shot scenarios, respectively.

The main contributions of the paper are summarised as follows:
\begin{itemize}
  \item We introduce feature registration as a category-agnostic approach for few-shot anomaly detection (FSAD). To our best of knowledge, it is the first FSAD method that trains a single generalizable model and requires no re-training or parameter fine-tuning for new categories.
  \item Extensive experiments on recent benchmark datasets have shown that the proposed RegAD outperforms the state-of-the-art FSAD methods on both the anomaly detection and anomaly localization tasks.
\end{itemize}

\section{Related Work}
\subsection{Anomaly Detection} 
AD is a task where training datasets contain only normal data. To better estimate the normal distributions, one-class classification based approaches tend to depict the normal data directly with statistical approaches~\cite{Eskin2000Anomaly,scholkopf2001estimating,Rahmani2017Coherence,ruff2018deep}. Self-supervised based approaches are trained using only normal data, and then make inferences by assuming that anomalous data performs differently. In this domain, reconstruction~\cite{xia2015learning,schlegl2017unsupervised,zong2018deep,Sabokrou2018Adversarially,ganomaly,gong2019memorizing,metaformer,huang2022ssm} is the most popular self-supervision. Some approaches~\cite{golan2018deep,ARNet,MKD} introduce other self-supervisions, \emph{e.g.}, \cite{golan2018deep} applies dozens of image geometric transforms for transformation classification; \cite{ARNet} proposes a restoration framework for attribute restoration. Recent AD methods usually use feature embeddings extracted from a pre-trained deep neural network. Feature embedding is mostly used as an input for a traditional machine learning algorithm or statistical metrics such as the Mahalanobis distance~\cite{defard2021padim}. The network used as a feature extractor can be trained from scratch~\cite{yi2020patch}, while several methods~\cite{cutpaste,defard2021padim,focus,patchcore,cflow} have also achieved state-of-the-art results using models pre-trained on the ImageNet dataset~\cite{imagenet}. This paper differs from these previous works by focusing on FSAD, where only a few normal images are available.

\subsection{Few-shot Learning}
Few-shot learning (FSL) aims to adapt to novel classes with a few annotated examples. Representative FSL methods can be categorized into metric learning, generation, and optimization. Metric learning approaches~\cite{snell2017prototypical,sung2018learning,he2021revisiting} learn to calculate a feature space that classifies an unseen sample based on its nearest example category. Generation methods \cite{liu2020deep,yang2021free,chen2019multi} enhance the novel class performance by generating its images or features. Optimization methods \cite{Ravi2017OptimizationAA,finn2017model} learn commonalities among different categories and explore efficient optimization strategies for novel classes based on these commonalities. In this paper, the proposed method predicts `normal' or `anomaly' for a new category. In contrast to previous work on FSL, both training data and support set only have positive (normal) examples without any negative (anomaly) samples. 

\subsection{Few-shot Anomaly Detection}
FSAD aims to indicate anomalies with only a few normal samples as the support images for target categories. TDG~\cite{TDG} proposes a hierarchical generative model that captures the multi-scale patch distribution of each support image. They use multiple image transformations and optimize discriminators to distinguish between real and fake patches, as well as between different transformations applied to the patches. The anomaly score was obtained by aggregating the patch-based votes of the correct transformations.
DiffNet~\cite{DiffNet} leverages the descriptiveness of features extracted by convolutional neural networks to estimate their density using a normalizing flow, which is a tool well-suited to estimate distributions from a few support samples. Metaformer~\cite{metaformer} can be applied to the FSAD, although an additional large-scale dataset, MSRA10K~\cite{msra10k}, should be used during its entire meta-training procedure (beyond parameter pre-training), together with additional pixel-level annotations. In this paper, we design registration based FSAD to learn the category-agnostic feature registration, enabling the model to detect anomalies in new categories given a few normal images without fine-tuning. 

\begin{figure}[t]
\centering
\includegraphics[width=1.0\textwidth]{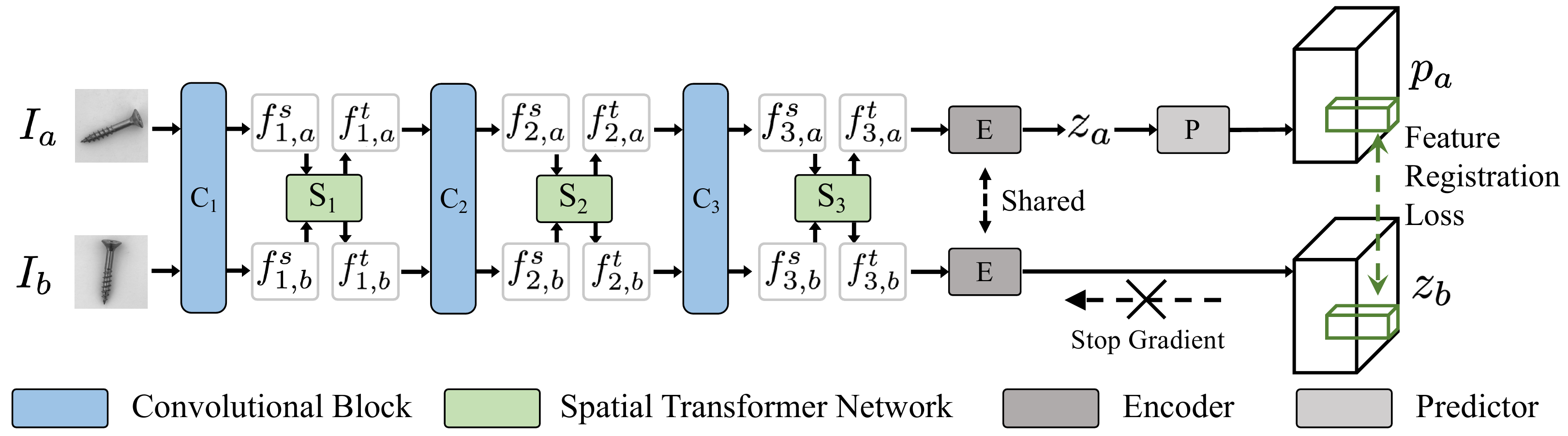}
\caption{The model architecture of the proposed RegAD. Given paired images from the same category, features are extracted by three convolutional residual blocks each followed by a spatial transformer network. A Siamese network acts as the feature encoder, supervised by a registration loss for feature similarity maximization.}
\label{img:RegAD}
\end{figure}

\section{Problem Setting}
\label{sec:PS}
We first formally define the problem setting for the proposed few-shot anomaly detection. Given a training set consisting of only normal samples of $n$ categories, \emph{i.e.,} $\mathcal{T}_{train}=\bigcup_{i=1}^{n}\mathcal{T}_{i}$, where the subset $\mathcal{T}_i$ consists of normal samples from the category $c_i$, ($i=1,2,\cdots,n$), we want to train a category-agnostic anomaly detection model. At test time, given a normal or anomalous image from a target category $c_t\ (t \notin \{1,2,\cdots,n\})$ and its associated support set $\mathcal{S}_t$ consisting of $k$ normal samples from the target category $c_t$, the trained category-agnostic anomaly detection model should predict whether the image is anomalous or not.

For FSAD, we attempt to detect anomalies from test samples of unseen/novel categories using only a few normal images as the support set. The key challenges lie in: (i) $\mathcal{T}_{train}$ has only access to normal samples from multiple known categories (\emph{e.g.}, different objects or textures), without any image-level or pixel-level annotations, (ii) the test data is from an unseen/novel category, and (iii) only a few normal samples from the target category $c_t$ are available, making it hard to estimate the normal distribution of the target category $c_t$.

\section{Method}\label{sec:method}
Motivated by how humans detect anomalies, the feature registration is used as a generalization paradigm for FSAD. During the training procedure, we leverage an anomaly-free feature registration network to learn category-agnostic feature registration. During testing, given the support set of a few normal images, the normal distribution of registered features for the target category is estimated with a statistical-based distribution estimator. Test samples that are out of the learned statistical normal distribution are considered anomalies.

\subsection{Feature Registration Network}
Given a pair of images $I_a$ and $I_b$ randomly selected from a same category in the training set $\mathcal{T}_{train}$, a ResNet-type convolutional network~\cite{he2016deep} is leveraged as the feature extractor. Specifically, as shown in Fig.~\ref{img:RegAD},  the first three convolutional residual blocks of ResNet, $C_1$, $C_2$, and $C_3$, are adopted, and the last convolution block in ResNet's original design is discarded, in order to ensure that final features still retain spatial information. A spatial transformer network (STN)~\cite{STN} is inserted into each block as a feature transformation module, so as to enable the model to learn feature registration flexibly, inspired by~\cite{focus}. Specifically, a transformation function $S_i$ ($i=1,2,3$) is applied on an input feature $f_{i}^s$:
\begin{equation}
\begin{pmatrix}
x_i^t \\ y_i^t
\end{pmatrix}
= S_i(f_{i}^s)
=A_i\begin{pmatrix}
x_{i}^s \\ y_{i}^s \\ 1
\end{pmatrix}
=\begin{bmatrix}
\theta_{11} & \theta_{12} & \theta_{13} \\
\theta_{21} & \theta_{22} & \theta_{23} \\
\end{bmatrix} 
\begin{pmatrix}
x_{i}^s \\ y_{i}^s \\ 1
\end{pmatrix},
\end{equation}
where $(x_i^t,y_i^t)$ are the target coordinates of output feature $f_i^t$, $(x_i^s,y_i^s)$ are the same points in the source coordinates of input feature $f_i^s$ and $A_i$ is the affine transformation matrix. The module $S_i$ is used to learn the mappings from features of convolutional block $C_i$ with the same tiny architecture as used in~\cite{STN}. 

Given paired extracted features $f_{3,a}^t$ and $f_{3,b}^t$ as the final transformation outputs, we design the feature encoder as a Siamese network~\cite{bromley1993signature}. A Siamese network is a parameter-sharing neural network applied on multiple inputs. To avoid the collapsing problem when optimized without negative pairs, inspired by SimSiam~\cite{chen2021exploring}, features are processed by the same encoder network $E$ followed by a prediction head $P$ applied on one branch. A stop-gradient operation is applied on the other branch, as shown in Fig.~\ref{img:RegAD}, which is critical to prevent such collapsing solutions. Denote $p_a\triangleq P(E(f_{3,a}))$ and $z_b\triangleq E(f_{3,b})$, a negative cosine similarity loss is applied:
\begin{equation}
    \mathcal{D}(p_a,z_b)=-\frac{p_a}{||p_a||_2}\cdot \frac{z_b}{||z_b||_2},
\end{equation}
where $||\cdot||_2$ is a $L_2$ norm. 
Instead of registering the images pixel-by-pixel, here we use a feature-level registration loss which may be deemed as a relaxed version of the pixel-wise registration constraints for better robustness. Finally, following SimSiam~\cite{chen2021exploring}, a symmetrized feature registration loss is defined as:
\begin{equation}\label{eq:loss}
    \mathcal{L}=\frac{1}{2}(\mathcal{D}(p_a,z_b)+\mathcal{D}(p_b,z_a)).
\end{equation}

\textbf{Discussion.}  
Features from the proposed method retain relatively complete spatial information, since we adopt the first three convolutional blocks of ResNet as the backbone without global average pooling, followed by a convolutional encoder and predictor architecture, but not the MLP architecture in SimSiam~\cite{chen2021exploring}. Thus Eq.~\eqref{eq:loss} should be computed by averaging cosine similarity scores at every spatial pixel. Features containing spatial information are beneficial for the AD task, which needs to provide anomaly score maps as prediction results. Different from SimSiam~\cite{chen2021exploring}, which defines the inputs as two augmentations of one image and maximizes their similarity to enhance the model representation, the proposed feature registration leverages two different images as inputs and maximizes the similarity between the features to learn the registration. 

\subsection{Normal Distribution Estimation}
To perform testing, it is assumed that the feature registration ability can generalize to the target category, and the learned feature registration model is applied to the support set $\mathcal{S}_t$ for the target category without parameter fine-tuning. Multiple data augmentations are applied to the support images, consistent with~\cite{TDG}. As the two branches of the Siamese network are exactly the same, only one branch feature is used for the normal distribution estimation. After achieving the registered features, a statistical-based estimator~\cite{defard2021padim} is used to estimate the normal distribution of target category features, which uses multivariate Gaussian distributions to get a probabilistic representation of the normal class. Suppose an image is divided into a grid of $(i,j)\in [1,W]\times [1,H]$ positions where $W\times H$ is the resolution of features used to estimate the normal distribution. At each patch position $(i,j)$, let $F_{ij} = \{ f_{ij}^k,k\in [1,N]\}$ be the registered features from $N$ augmented support images. $f_{ij}$ is the aggregated features at patch position $(i,j)$, achieved by concatenating the three STN outputs at the corresponding position with upsampling operations to match their sizes. By the assumption that $F_{ij}$ is generated by $\mathcal{N}(\mu_{ij}, \Sigma_{i j})$, the sample covariance is:
\begin{equation}
    \Sigma_{i j}=\frac{1}{N-1} \sum_{k=1}^{N}\left(f_{ij}^k-\mu_{i j}\right)\left(f_{ij}^k-\mu_{i j}\right)^{\mathrm{T}}+\epsilon I,
\end{equation}
where $\mu_{ij}$ is the sample mean of $F_{ij}$, and the regularization term $\epsilon I$ makes the sample covariance matrix full rank and invertible. Finally, each possible patch position is associated with a multivariate Gaussian distribution.

\textbf{Discussion.} Data augmentations are widely adopted in AD, and especially in FSAD, including TDG~\cite{TDG} and DiffNet~\cite{DiffNet}. However, most methods simply apply the data augmentations on both the support and test images without any exploration of the impact. In this paper, we emphasize that data augmentation plays a very important role in expanding the support set, which is beneficial for the normal distribution estimation. Specifically, we adopt augmentations including rotation, translation, flipping, and graying for each image in the support set $\mathcal{S}_t$. Other augmentations like mixup and cutpaste are not considered since they seem more suitable for simulating anomalies~\cite{cutpaste}. We conduct the possible combinations of all these augmentations for each sample in the support set, which jointly combine into a larger support set. We conduct the normal distribution estimation on such an augmented support set. We study the impacts of different augmentations in the supplementary material.

\subsection{Inference} 
During inference, test samples that are out of the normal distribution are considered anomalies. For each test image in $\mathcal{T}_{test}$, we use the Mahalanobis distance $\mathcal{M}\left(f_{i j}\right)$ to give an anomaly score to the patch in position $(i,j)$, where
\begin{equation}
    \mathcal{M}\left(f_{i j}\right)=\sqrt{\left(f_{i j}-\mu_{i j}\right)^{T} \Sigma_{i j}^{-1}\left(f_{i j}-\mu_{i j}\right)}.
\end{equation}
The matrix of Mahalanobis distances $\mathcal{M}=\left(\mathcal{M}\left(f_{i j}\right)\right)_{1\leqslant i\leqslant W, 1\leqslant j\leqslant H}$ forms an anomaly map. Three inverse affine transformations corresponding to the three STN modules are applied to this anomaly map to get the final anomaly score map $\mathcal{M}_{final}$ aligned with the original image. High scores in this map indicate the anomalous areas. The final anomaly score of the entire image is the maximum of anomaly map $\mathcal{M}_{final}$. Compared with~\cite{TDG,DiffNet}, RegAD cancels the data augmentation of the test images which reduces the inference computational costs.

\section{Experiments}
\subsection{Experimental Setups}
\subsubsection{Datasets.}
We experiment on two challenging real-world benchmark datasets for AD~\cite{bergmann2019mvtec,jezek2021deep}, which are both related to industrial defect detection.
\begin{itemize}
    \item \textbf{MVTec}~\cite{bergmann2019mvtec}: MVTec comprises 15 categories with 3629 images for training and validation and 1725 images for testing. The training set contains only of normal images without defects. The test set contains both images with various kinds of defects (anomaly) and defect-free images (normal). On average five per category, 73 different defect types are given. 
    All images are in the resolution range between $700 \times 700$ and $1024 \times 1024$ pixels. Pixel-wise ground truth labels for each defective image region are provided. 
    \item \textbf{MPDD}~\cite{jezek2021deep}: MPDD is a newly proposed dataset focused specifically on defect detection during painted metal part fabrication, containing 6 classes of metal parts. Images are captured under the conditions of various spatial orientations, positions, and distances of multiple objects, concerning different light intensities and a non-homogeneous background. 
\end{itemize}

For each dataset, we conduct experiments on two different experimental settings. (i) \textbf{Aggregated training} on multiple categories and then adapting to unseen categories, and (ii) \textbf{Individual training} only with the support set for each category.

\subsubsection{Competing Methods.} 
We consider two state-of-the-art FSAD approaches, TDG~\cite{TDG} and DiffNet~\cite{DiffNet}. These two methods both train models individually for each category (setting (ii)). Results are reproduced using the official source code. Considering that our method uses data from multiple categories, for fairness of comparison, we extend them to leverage the same amount of data (setting (i)). A pre-training procedure is added to these methods, where data from multiple categories are used to pre-train the transformation classifier for TDG or initialize the normalizing flow-based estimator for DiffNet. The corresponding methods are TDG+ and DiffNet+. We also evaluate RegAD under the individual training setting, and denote the corresponding method as RegAD-L. We compare with some state-of-the-art vanilla AD methods, such as GANomaly~\cite{ganomaly}, ARNet~\cite{ARNet}, MKD~\cite{MKD}, CutPaste~\cite{cutpaste}, FYD~\cite{focus}, PaDiM~\cite{defard2021padim}, PatchCore~\cite{patchcore} and CflowAD \cite{cflow}. These methods use the whole normal dataset for their training, so they can be deemed as the upper bound on FSAD performance.

\subsubsection{Evaluation Protocols.} 
We quantify the model performance using the area under the Receiver Operating Characteristic (ROC) curve metric (AUC), which is commonly adopted as the performance measurement for AD tasks. The image-level AUC and the pixel-level AUC are used for anomaly detection and anomaly localization respectively.

\subsubsection{Model Configuration and Training Details.}
An ImageNet pre-trained ResNet-18~\cite{he2016deep} is used as the backbone, followed by a convolutional-based encoder and predictor. To retain the spatial information, the encoder contains three $1\times 1$ convolutional layers, while the predictor contains two $1\times 1$ convolutional layers, without any pooling operation. We train models on $224 \times 224$ images on one NVIDIA GTX 3090. We update the parameters using momentum SGD with a learning rate of 0.0001 for 50 epochs, with a batch size of 32. A single cycle of cosine learning rate is used as the decay schedule. 

\begin{table}[t]
\centering
\caption{Results of k-shot anomaly detection on the MVTec dataset, comparing with state-of-the-art methods. Results are listed as the average AUC in \% of 10 runs and are marked individually for each category. A macro-average score over all categories is also reported in the last row. The best-performing method is in bold.}
\label{tal:mvtec}
\scriptsize
\setlength{\tabcolsep}{1.2pt}{
\begin{tabular}{C{1.3cm}C{1.1cm}C{1.2cm}C{0.9cm}C{1.1cm}C{1.2cm}C{0.9cm}C{1.1cm}C{1.2cm}C{0.9cm}}
\toprule
\multirow{3}{*}{Category} & \multicolumn{3}{c}{k=2} & \multicolumn{3}{c}{k=4} & \multicolumn{3}{c}{k=8}\\
\cmidrule(lr){2-4} \cmidrule(lr){5-7} \cmidrule(lr){8-10}
& \makecell[c]{TDG+\\\cite{TDG}} & \makecell[c]{DiffNet+\\\cite{DiffNet}} & \makecell[c]{RegAD\\(ours)} & \makecell[c]{TDG+\\\cite{TDG} } & \makecell[c]{DiffNet+\\\cite{DiffNet}} & \makecell[c]{RegAD\\(ours)} & \makecell[c]{TDG+\\\cite{TDG}} & \makecell[c]{DiffNet+\\\cite{DiffNet} } & \makecell[c]{RegAD\\(ours)}\\
\cmidrule(lr){1-1} \cmidrule(lr){2-2} \cmidrule(lr){3-3} \cmidrule(lr){4-4} \cmidrule(lr){5-5} \cmidrule(lr){6-6} \cmidrule(lr){7-7} \cmidrule(lr){8-8} \cmidrule(lr){9-9} \cmidrule(lr){10-10}
Bottle     & 69.3 & 99.3 & \textbf{99.4} & 69.6 & 99.3 & \textbf{99.4} & 70.3 & 99.4 & \textbf{99.8} \\
Cable      & 68.3 & \textbf{85.3} & 65.1 & 70.3 & \textbf{85.2} & 76.1 & 74.7 & \textbf{87.9} & 80.6 \\
Capsule    & 55.1 & \textbf{73.0} & 67.5 & 47.6 & \textbf{80.3} & 72.4 & 44.7 & \textbf{78.6} & 76.3 \\
Carpet     & 66.2 & 78.4 & \textbf{96.5} & 68.7 & 78.6 & \textbf{97.9} & 78.2 & 78.5 & \textbf{98.5} \\
Grid       & 83.8 & 62.1 & \textbf{84.0} & 86.2 & 60.5 & \textbf{91.2} & 87.6 & 78.5 & \textbf{91.5} \\
Hazelnut   & 67.2 & 94.9 & \textbf{96.0} & 71.2 & \textbf{95.8} & 95.8 & 82.8 & \textbf{97.9} & 96.5 \\
Leather    & 93.6 & 90.7 & \textbf{99.4}  & 93.2 & 91.2 & \textbf{100} & 93.5 & 92.2 & \textbf{100}  \\
Metal Nut  & 67.1 & 61.9 & \textbf{91.4} & 69.2 & 67.3 & \textbf{94.6} & 68.7 & 67.6 & \textbf{98.3} \\
Pill       & 69.2 & \textbf{83.2} & 81.3 & 64.7 & \textbf{84.0} & 80.8 & 67.9 & \textbf{82.1} & 80.6 \\
Screw      & \textbf{98.8} & 73.4 & 52.5 & \textbf{98.8} & 72.5 & 56.6 & \textbf{99.0} & 75.0 & 63.4 \\
Tile       & 86.3 & \textbf{97.0} & 94.3 & 87.2 & \textbf{98.0} & 95.5 & 87.4 & \textbf{99.6} & 97.4 \\
Toothbrush & 54.4 & 60.8 & \textbf{86.6} & 57.8 & 62.5 & \textbf{90.9} & 57.6 & 60.8 & \textbf{98.5} \\
Transistor & 55.9 & 61.8 & \textbf{86.0} & 67.7 & 62.2 & \textbf{85.2} & 71.5 & 63.3 & \textbf{93.4} \\
Wood       & 98.4 & 98.1 & \textbf{99.2} & 98.3 & 96.4 & \textbf{98.6} & 98.4 & \textbf{99.4} & \textbf{99.4} \\
Zipper     & 64.4 & \textbf{89.2} & 86.3 & 65.3 & 84.8 & \textbf{88.5} & 66.3 & 87.3 & \textbf{94.0} \\
\cmidrule(lr){1-1} \cmidrule(lr){2-2} \cmidrule(lr){3-3} \cmidrule(lr){4-4} \cmidrule(lr){5-5} \cmidrule(lr){6-6} \cmidrule(lr){7-7} \cmidrule(lr){8-8} \cmidrule(lr){9-9} \cmidrule(lr){10-10}
Average    & 73.2 & 80.6 & \textbf{85.7} & 74.4 & 81.3 & \textbf{88.2} & 76.6 & 83.2 & \textbf{91.2} \\
\bottomrule
\end{tabular}}
\end{table}

\begin{table}[t]
\centering
\caption{Results of k-shot anomaly detection on the MPDD dataset, comparing with state-of-the-art methods. Results are listed as the average AUC in \% of 10 runs and are marked individually for each category. A macro-average score over all categories is also reported in the last row. The best-performing method is in bold.}
\label{tal:mpdd}
\scriptsize
\setlength{\tabcolsep}{0.8pt}{
\begin{tabular}{C{1.8cm}C{1.1cm}C{1.2cm}C{0.9cm}C{1.1cm}C{1.2cm}C{0.9cm}C{1.1cm}C{1.2cm}C{0.9cm}}
\toprule
\multirow{3}{*}{Category} & \multicolumn{3}{c}{k=2} & \multicolumn{3}{c}{k=4} & \multicolumn{3}{c}{k=8}\\
\cmidrule(lr){2-4} \cmidrule(lr){5-7} \cmidrule(lr){8-10}
& \makecell[c]{TDG+\\\cite{TDG} } & \makecell[c]{DiffNet+\\\cite{DiffNet} } & \makecell[c]{RegAD\\(ours)} & \makecell[c]{TDG+\\\cite{TDG} } & \makecell[c]{DiffNet+\\\cite{DiffNet} } & \makecell[c]{RegAD\\(ours)} & \makecell[c]{TDG+\\\cite{TDG} } & \makecell[c]{DiffNet+\\\cite{DiffNet} } & \makecell[c]{RegAD\\(ours)}\\
\cmidrule(lr){1-1} \cmidrule(lr){2-2} \cmidrule(lr){3-3} \cmidrule(lr){4-4} \cmidrule(lr){5-5} \cmidrule(lr){6-6} \cmidrule(lr){7-7} \cmidrule(lr){8-8} \cmidrule(lr){9-9} \cmidrule(lr){10-10}
bracket black & 46.4 & 56.7 & \textbf{63.3} & 48.8 & 59.9 & \textbf{63.8} & 51.0 & \textbf{69.7} & 67.3 \\
bracket brown & 54.9 & \textbf{61.3} & 59.4 & 57.5 & 64.2 & \textbf{66.1} & 65.4 & 66.3 & \textbf{69.6} \\
bracket white & \textbf{64.0} & 42.2 & 55.6 & \textbf{65.4} & 51.8 & 59.3 & 66.8 & \textbf{69.1} & 61.4 \\
connector     & 53.1 & 54.1 & \textbf{73.0} & 55.8 & 54.8 & \textbf{77.2} & 62.9 & 54.5 & \textbf{84.9} \\
metal plate   & 91.8 & \textbf{96.8} & 61.7 & 95.1 & \textbf{98.2} & 78.6 & 98.4 & \textbf{98.8} & 80.2 \\
tubes         & 51.8 & 49.8 & \textbf{67.1} & 58.5 & 50.7 & \textbf{67.5} & 64.9 & 52.6 & \textbf{67.9} \\
\cmidrule(lr){1-1} \cmidrule(lr){2-2} \cmidrule(lr){3-3} \cmidrule(lr){4-4} \cmidrule(lr){5-5} \cmidrule(lr){6-6} \cmidrule(lr){7-7} \cmidrule(lr){8-8} \cmidrule(lr){9-9} \cmidrule(lr){10-10}
Average       & 60.3 & 60.2 & \textbf{63.4} & 63.5 & 63.3 & \textbf{68.3} & 68.2 & 68.5 & \textbf{71.9} \\
\bottomrule
\end{tabular}}
\end{table}

\begin{table}[t]
\centering
\caption{Results of anomaly detection on the MVTec and MPDD datasets under two different experimental settings (i) and (ii), comparing with state-of-the-art few-shot anomaly detection methods on $k=2, 4, 8$. Results are listed as the macro-average AUC in \% over all categories in each dataset of 10 runs. The best-performing method for each experimental setting is in bold.}
\label{tal:fewshot}
\scriptsize
\setlength{\tabcolsep}{0.6pt}{
\begin{tabular}{C{2.2cm}C{1.3cm}C{1.5cm}C{1.8cm}C{0.8cm}C{0.8cm}C{0.8cm}C{0.8cm}C{0.8cm}C{0.8cm}}
\toprule
\multicolumn{1}{c}{\multirow{2}{*}{Methods}}
& ImageNet & Aggregated & Time of & \multicolumn{3}{c}{MVTec} & \multicolumn{3}{c}{MPDD} \\
& Pretrain & Training &  Adaptation & k=2 & k=4 & k=8 & k=2 & k=4 & k=8\\
\cmidrule(lr){1-1} \cmidrule(lr){2-2} \cmidrule(lr){3-3} \cmidrule(lr){4-4} \cmidrule(lr){5-5} \cmidrule(lr){6-6} \cmidrule(lr){7-7} \cmidrule(lr){8-8} \cmidrule(lr){9-9} \cmidrule(lr){10-10}
TDG~\cite{TDG} & \checkmark & \ding{55} & - & 71.2 & 72.7 & 75.2 & 57.3 & 60.4 & 64.4\\
DiffNet~\cite{DiffNet} & \checkmark & \ding{55} & - & 80.5 & 80.8 & 82.9 & \textbf{58.4} & \textbf{61.2} & \textbf{66.5}\\
RegAD-L (ours) & \checkmark & \ding{55} & - & \textbf{81.5} & \textbf{84.9} & \textbf{87.4} & 50.8 & 54.2 & 61.1\\
\cmidrule(lr){1-1} \cmidrule(lr){2-2} \cmidrule(lr){3-3} \cmidrule(lr){4-4} \cmidrule(lr){5-5} \cmidrule(lr){6-6} \cmidrule(lr){7-7} \cmidrule(lr){8-8} \cmidrule(lr){9-9} \cmidrule(lr){10-10}
TDG+~\cite{TDG}   & \checkmark & \checkmark & 1559.76s & 73.2 & 74.4 & 76.6 & 60.3 & 63.5 & 68.2\\
DiffNet+~\cite{DiffNet}   & \checkmark & \checkmark & 357.75s & 80.6 & 81.3 & 83.2 & 60.2 & 63.3 & 68.5\\
RegAD (ours) & \checkmark & \checkmark & 4.47s & \textbf{85.7} & \textbf{88.2} & \textbf{91.2} & \textbf{63.4} & \textbf{68.3} & \textbf{71.9}\\
\bottomrule
\end{tabular}}
\end{table}

\begin{table}[t]
\centering
\caption{Results of anomaly detection and anomaly localization on the MVTec and MPDD datasets, comparing with state-of-the-art vanilla AD methods. Results are listed as AUC in \% as the macro-average score over all categories in each dataset.}
\label{tal:vanilla}
\scriptsize
\setlength{\tabcolsep}{1.2pt}{
\begin{tabular}{C{2.2cm}C{1.5cm}C{1.6cm}C{1.8cm}C{1.1cm}C{1.1cm}C{1.1cm}C{1.1cm}}
\toprule
\multicolumn{1}{c}{\multirow{2}{*}{Methods}}
& \multicolumn{1}{c}{\multirow{2}{*}{Data}} & ImageNet & \multicolumn{1}{c}{\multirow{2}{*}{Backbone}} & \multicolumn{2}{c}{MVTec} & \multicolumn{2}{c}{MPDD} \\
& & Pretrain & & image & pixel & image & pixel\\
\cmidrule(lr){1-1} \cmidrule(lr){2-2} \cmidrule(lr){3-3} \cmidrule(lr){4-4} \cmidrule(lr){5-5} \cmidrule(lr){6-6} \cmidrule(lr){7-7} \cmidrule(lr){8-8}
RegAD (k=4) & 4 images & \checkmark & Res18 & 88.2 & 95.8 & 68.8 & 93.9\\
RegAD (k=8) & 8 images & \checkmark & Res18 & 91.2 & 96.7 & 71.9 & 95.1\\
RegAD (k=16) & 16 images & \checkmark & Res18 & 92.7 & 96.6 & 75.3 & 96.3\\
RegAD (k=32) & 32 images & \checkmark & Res18 & 94.6 & 96.9 & 76.8 & 96.3\\
\cmidrule(lr){1-1} \cmidrule(lr){2-2} \cmidrule(lr){3-3}\cmidrule(lr){4-4} \cmidrule(lr){5-6} \cmidrule(lr){7-8}
GANomaly~\cite{ganomaly} & full data & \ding{55} & UNet & 80.5 & - & 64.8 & -\\
ARNet~\cite{ARNet} & full data & \ding{55} & UNet & 83.9 & - & 69.7 & -\\
MKD~\cite{MKD} & full data & \checkmark & Res18 & 87.7 & 90.7 & - & - \\
CutPaste~\cite{cutpaste} & full data & \checkmark & Res18 & 95.2 & 96.0 & - & -\\
FYD~\cite{focus} & full data & \checkmark & Res18 & 97.3 & 97.4 & - & -\\
PaDiM~\cite{defard2021padim} & full data & \checkmark & WRN50 & 97.9 & 97.5 & 74.8 & 96.7\\
PatchCore~\cite{patchcore} & full data & \checkmark & WRN50 & 99.1 & 98.1 & 82.1 & 95.7\\
CflowAD~\cite{cflow} & full data & \checkmark & WRN50 & 98.3 & 98.6 & 86.1 & 97.7\\
\bottomrule
\end{tabular}}
\end{table}

\subsection{Comparison with State-of-the-art Methods}
\subsubsection{Comparison with Few-Shot Anomaly Detection Methods.} 
Experiments were conducted using the leave-one-out setting, \emph{i.e.}, a target category was chosen to be tested, while other categories in the dataset are used for training. Table~\ref{tal:mvtec} and Table~\ref{tal:mpdd} show the comparison results on MVTec and MPDD, respectively, under the experimental setting (i). RegAD achieves an improvement of 5.1\%, 6.9\%, 8.0\% in average AUC on MVTec, and an improvement of 3.2\%, 5.0\%, 3.4\% in average AUC on MPDD, over DiffNet+~\cite{DiffNet}, with 2-shot, 4-shot, and 8-shot scenarios, respectively. Also, with one-shot, RegAD achieves 82.4\% and 57.8\% AUC on MVtec and MPDD respectively.

RegAD is tested without any parameter fine-tuning, which may not guarantee the best performance for every category, while other baselines have unfair advantages in that they tune the parameters for each category. In 9 out of the 15 categories, RegAD outperforms all the other baselines. RegAD also achieves the least standard deviation (10.94) for the 15 categories when k=8, compared to TDG+ (15.20) and DiffNet+ (13.11), suggesting its better generalizability across different categories. Also, although using different training settings, for MVTec (k=8), RegAD achieves 91.2\% AUC, with an $\approx$3\% improvement compared with Metaformer~\cite{metaformer} which uses an additional large-scale dataset, MSRA10K~\cite{msra10k}, during its entire training procedure.

\textbf{Discussion.} Adaptation time is important for real-world applications of FSAD. The procedures of fine-tuning for both TDG+ and DiffNet+ are time-consuming since they update the models for many epochs, while RegAD has the fastest adaptation speed since it is based on a statistical estimator which needs only one inference for each support image. In Table~\ref{tal:fewshot}, we report the adaptation times for each method, by averaging the results for $k=2,4,8$ on both the MVTec and MPDD datasets. Compared with TDG+ (1559.76s) and DiffNet+ (357.75s), the proposed RegAD has the fastest adaptation speed (4.47s).

Table~\ref{tal:fewshot} also compares these methods under experimental setting (ii), where we train the models individually using the support images for each category. RegAD-L means RegAD with individual training on one category only. Assuming that features pre-trained by ImageNet are fully representative, we simply fine-tune features using limited support images. Thus, we conduct the fine-tuning procedures directly under an ImageNet pre-training backbone for all methods. All methods use the same ImageNet pre-training backbone to have a fair comparison. In this setting, RegAD-L outperforms both TDG and DiffNet on the MVTec dataset. DiffNet performs better than the proposed method on the MPDD dataset. However, compared with RegAD-L, the proposed RegAD improves a lot, showing the effectiveness of the proposed feature registration aggregated training procedure on multiple categories.

\subsubsection{Comparison with Vanilla Anomaly Detection Methods.} 
The state-of-the-art vanilla AD methods use the whole normal dataset for their training and train a separate model for each category, so their performance can be seen as the upper bound for FSAD. We consider methods including GANomaly~\cite{ganomaly}, ARNet~\cite{ARNet}, MKD~\cite{MKD}, CutPaste~\cite{cutpaste}, FYD~\cite{focus}, PaDiM~\cite{defard2021padim}, PatchCore~\cite{patchcore} and CflowAD~\cite{cflow}. Results in Table~\ref{tal:vanilla} show that the proposed RegAD reaches competitive performance even compared with vanilla AD methods that are based on extensive normal data. For example, with only 4 support images, the proposed method (88.2\% AUC) outperforms MKD (87.7\%) with the same ImageNet pre-trained backbone, and with 32 support images its AUC increases to 94.6\%.

\subsection{Ablation Studies}\label{sec:abl}
Experiments were performed to evaluate the contribution made by individual components of the proposed method. Results of ablation studies for k-shot anomaly detection and localization on the MVTec and MPDD datasets are shown in Table~\ref{tal:abl_all}. Modules of ‘A’, ‘F’, and ‘S’ mean the augmentations for support sets, the feature registration aggregated training on multiple categories, and the spatial transformer networks (STN), respectively. Results in Table~\ref{tal:abl_all} show that: 

\begin{table}[t]
\centering
\caption{Ablation studies of k-shot anomaly detection and localization on the MVTec and MPDD datasets. Modules of `A', `F', and `S' mean the augmentations for the support set, the feature registration aggregated training, and the spatial transformer networks (STN), respectively. Results are listed as the macro-average AUC in \% over all categories in each dataset of 10 runs. The best-performing method is in bold.}
\label{tal:abl_all}
\scriptsize
\setlength{\tabcolsep}{1.1pt}{
\begin{tabular}{C{0.7cm}C{0.7cm}C{0.7cm}C{0.7cm}C{0.7cm}C{0.7cm}C{0.7cm}C{0.7cm}C{0.7cm}C{0.7cm}C{0.7cm}C{0.7cm}C{0.7cm}C{0.7cm}C{0.7cm}}
\toprule
\multicolumn{3}{c}{\multirow{2}{*}{Modules}}
& \multicolumn{6}{c}{MVTec}                                        & \multicolumn{6}{c}{MPDD}                                          \\
\cmidrule(lr){4-9} \cmidrule(lr){10-15}
& & & \multicolumn{3}{c}{image} & \multicolumn{3}{c}{pixel} & \multicolumn{3}{c}{image} & \multicolumn{3}{c}{pixel}  \\
\cmidrule(lr){1-3} \cmidrule(lr){4-6} \cmidrule(lr){7-9} \cmidrule(lr){10-12} \cmidrule(lr){13-15}
 A & F & S & k=2 & k=4 & k=8               & k=2 & k=4 & k=8                  & k=2 & k=4 & k=8               & k=2 & k=4 & k=8                   \\
\cmidrule(lr){1-1} \cmidrule(lr){2-2} \cmidrule(lr){3-3}\cmidrule(lr){4-4} \cmidrule(lr){5-5} \cmidrule(lr){6-6} \cmidrule(lr){7-7} \cmidrule(lr){8-8} \cmidrule(lr){9-9} \cmidrule(lr){10-10} \cmidrule(lr){11-11} \cmidrule(lr){12-12} \cmidrule(lr){13-13} \cmidrule(lr){14-14} \cmidrule(lr){15-15}
            &            &            & 74.7 & 78.0 & 80.5 & 88.6 & 90.5 & 92.1 & 49.6 & 53.7 & 55.5 & 89.5 & 91.2 & 92.0 \\
 \checkmark &            &            & 81.5 & 84.9 & 87.4 & 93.3 & 94.7 & 95.5 & 50.8 & 54.2 & 61.1 & 92.4 & 93.3 & 93.9\\
            & \checkmark &            & 78.0 & 80.9 & 83.1 & 90.8 & 92.5 & 94.0 & 53.9 & 55.5 & 57.2 & 91.5 & 92.2 & 93.0 \\
            & \checkmark & \checkmark & 79.1 & 82.9 & 84.9 & 90.5 & 93.3 & 94.3 & 57.6 & 60.9 & 62.7 & 91.0 & 91.8 & 93.0\\
 \checkmark & \checkmark &            & 83.0 & 86.4 & 89.3 & \textbf{94.7} & \textbf{95.9} & 96.6 & 52.8 & 57.7 & 64.8 & \textbf{93.3} & \textbf{94.1} & 94.4\\
 \checkmark & \checkmark & \checkmark & \textbf{85.7} & \textbf{88.2} & \textbf{91.2} & 94.6 & 95.8 & \textbf{96.7} & \textbf{63.4} & \textbf{68.8} & \textbf{71.9} & 93.2 & 93.9 & \textbf{95.1} \\
\bottomrule
\end{tabular}}
\end{table}

\begin{table}[t]
\centering
\caption{Ablation studies of different transformation versions of STN modules on MVTec and MPDD for anomaly detection with $k=2$. T, R means translation, and rotation, respectively. Results are listed as the macro-average AUC in \% over all categories in each dataset of 10 runs. The best-performing method is in bold.}
\label{tal:abl_stn}
\scriptsize
\setlength{\tabcolsep}{0.4pt}{
\begin{tabular}{C{1.3cm}C{1.3cm}C{0.8cm}C{0.8cm}C{0.9cm}C{0.9cm}C{1.1cm}C{1.4cm}C{1.1cm}C{1.3cm}C{1.0cm}}
\toprule
Data & no STN & T & R & scale & shear & \makecell[c]{R\\+scale} & \makecell[c]{T\\+scale} & \makecell[c]{T+R} & \makecell[c]{T+R\\+scale} & affine \\
\cmidrule(lr){1-1} \cmidrule(lr){2-2} \cmidrule(lr){3-3} \cmidrule(lr){4-4} \cmidrule(lr){5-5} \cmidrule(lr){6-6} \cmidrule(lr){7-7} \cmidrule(lr){8-8} \cmidrule(lr){9-9} \cmidrule(lr){10-10} \cmidrule(lr){11-11}
MVTec & 83.0 & 84.5 & 85.0 & 84.9 & 84.9 & \textbf{85.7} & 84.9 & 84.2 & 84.9 & 84.5 \\
MPDD & 52.8 & 62.3 & 57.7 & 59.2 & 59.0 & 61.5 & 61.8 & 61.0 & 61.7 & \textbf{63.4} \\
\bottomrule
\end{tabular}}
\end{table}

\begin{figure}[t]
\centering
\includegraphics[width=0.75\textwidth]{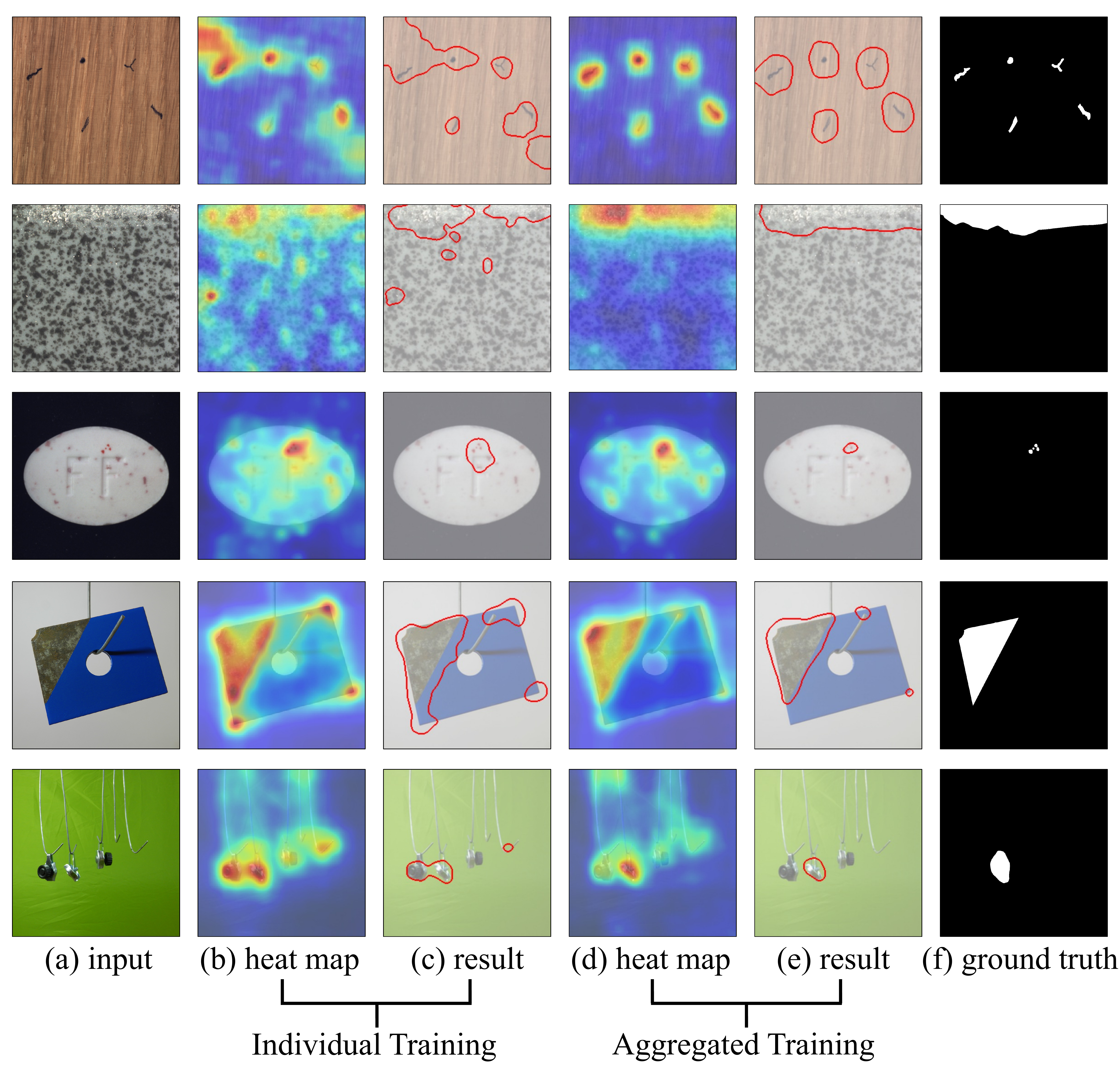}
\caption{Qualitative results of anomaly localization for RegAD on the MVTec dataset (top three rows) and the MPDD dataset (bottom two rows) for several cases, including localization results with individual training and aggregated training. Results from (e) show better performance than results from (c), showing the effectiveness of the proposed feature registration aggregated training procedure.}
\label{img:result}
\end{figure}

\begin{figure}[t]
  \begin{minipage}[t]{0.5\textwidth}
\centering
\includegraphics[width=5.0cm]{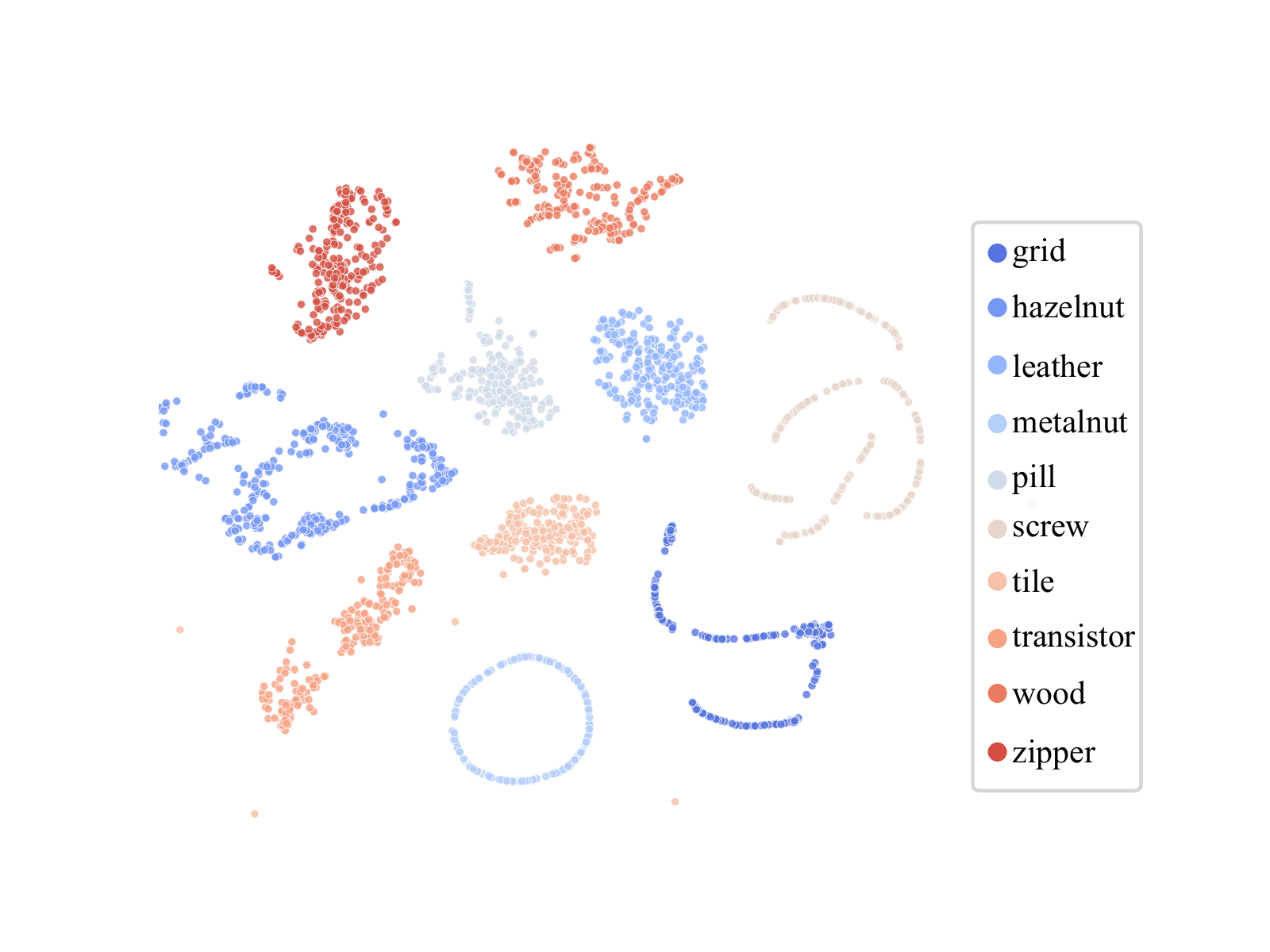}
\footnotesize

(a) Without Feature Registration
\end{minipage}
\begin{minipage}[t]{0.37\textwidth}
\centering
\includegraphics[width=5.0cm]{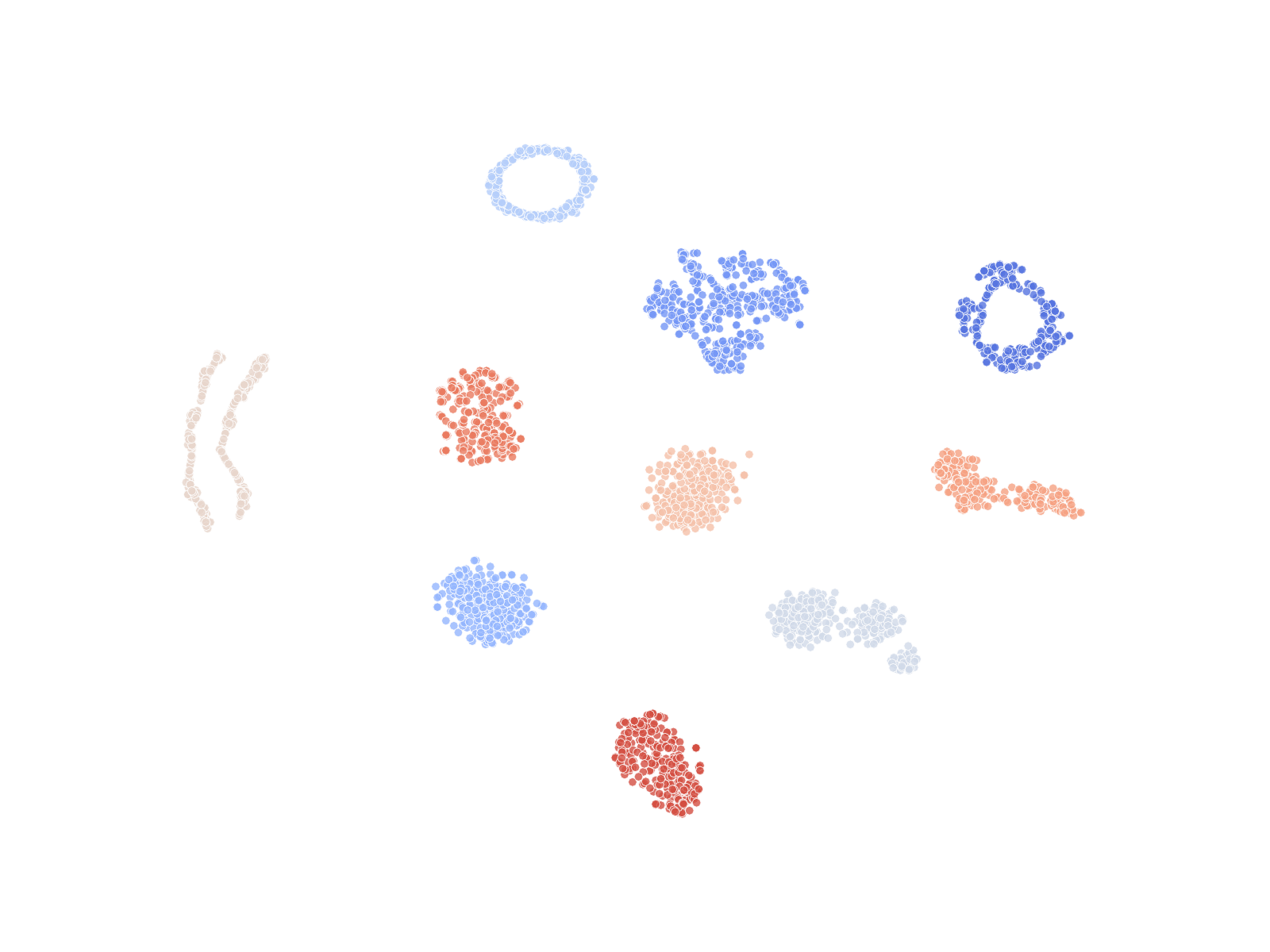}
\footnotesize
(b) With Feature Registration
\end{minipage}
\caption{Visualization, using t-SNE, of the features learned from the MVTec dataset, using (a) the baseline without the feature registration, and (b) the proposed method with the feature registration. The same t-SNE optimization iterations are used in each case. Results show that features with registration are more compact within each category, and more separated from different categories.}
\label{img:tsne}
\end{figure}

\textbf{(i) Augmentations.} The proposed support set augmentations are shown to be essential for both detection and localization. With $k=\{2,4,8\}$, the AUC is improved for 6.8\%, 6.9\%, 6.9\% on MVTec and for 1.2\%, 0.5\%, 0.6\% on MPDD, respectively. We further presents the ablation studies of comparing different augmentation methods for support images in the supplementary material.

\textbf{(ii) Feature Registration Aggregated Training.} The feature registration aggregated training on multiple categories is effective both with and without support image augmentations. It shows that the proposed feature registration is beneficial for estimating the normal distribution. As shown in Table~\ref{tal:abl_all}, with $k=\{2,4,8\}$, the proposed anomaly-free feature registration can improve the AUC by 3.3\%, 2.9\%, 2.6\% on MVTec, respectively.

\textbf{(iii) Spatial Transformer Modules.} The proposed STN module is good for improving the ability of the feature registration and thus beneficial for AD. For example, as shown in Table~\ref{tal:abl_all}, when $k=8$, the STN module can further improve the performance from 89.3\% to 91.2\% on MVTec and from 64.8\% to 71.9\% on MPDD. However, models with STN modules show similar pixel-level localization performance with models without STN modules. The reason comes from the information lost of the inverse transformation operation and its imprecision. These inverse transformations are designed as post-processing operations to rematch the spatial location of transformed features and the original images.

We further conduct ablation studies on different transformation versions of STN modules on MVTec and MPDD for AD, as shown in Table~\ref{tal:abl_stn}. The best performing STN version is rotation+scale on MVTec, which matches the observation that samples in this dataset are all aligned to the center, and thus, there is no need for translation. While for the MPDD dataset, since the samples are not well be centered, the version of STN with affine transformations shows the best performance. STN is used as a feature transformation module, enabling the model to implicitly transform the images to facilitate feature registration. Images in MPDD are captured under various spatial orientations and positions, thus aligning the features is expected to be helpful. For MVTec, objects are well centralized and have similar orientations, so STN is less helpful to MVTec.

\subsection{Visualization Analysis}
To qualitatively analyze how the proposed feature registration approach improves the anomaly localization performance, we visualize the results of some cases from the MVTec and MPDD datasets. It can be seen from the results in Fig.~\ref{img:result} that the localization produced by RegAD using aggregated training (column e) is closer to the ground truth (column f) than that produced by the individual training baseline (column c). This illustrates the effectiveness of the proposed feature registration training procedure on multiple categories.

We also use t-SNE~\cite{maaten2008visualizing} to visualize the features learned on the MVTec dataset, as shown in Fig.~\ref{img:tsne}. Each dot here represents an augmented normal sample from the test set. It can be seen that the proposed feature registration makes the features more compact within each category, and pushes away features of different categories, which is desirable for the benefit of estimating the normal distribution for each category.

\section{Conclusion}
This paper proposes an FSAD method utilizing registration, a task inherently generalizable across categories, as the proxy task. Given only a few normal samples for each category, we trained a category-agnostic feature registration network with the aggregated data. This model is shown to be directly generalizable to new categories, requiring no re-training or parameter fine-tuning. The anomalies are identified by comparing the registered features of the test image and its corresponding support (normal) images.  For both anomaly detection and anomaly localization, the method is shown to be competitive, even compared with vanilla AD methods that are trained with much larger volumes of data. The impressive results suggest a high potential for the proposed method to be applicable in real-world anomaly detection environments.

\textbf{Acknowledgments.}
This work is supported by the National Key Research and Development Program of China (No. 2020YFB1406801), 111 plan (No. BP0719010),  and STCSM (No. 18DZ2270700), and State Key Laboratory of UHD Video and Audio Production and Presentation.

\bibliographystyle{splncs04}
\bibliography{RegAD}

\appendix

\section{Main Contributions}

This paper targets a challenging yet practical setting for anomaly detection, with 1) a single model for all categories (\emph{i.e., generalizable without fine-tuning}), 2) only a few images for each novel category (\emph{i.e., few shot}), and 3) only normal samples available (\emph{i.e., unsupervised setting}). To our best knowledge, it is the first attempt to explore such a setting, as a critical step toward practical large-scale industrial applications, a point appreciated by the other reviewers. To learn a category-agnostic model, we further propose a novel comparison-based solution, which is quite different from the popular reconstruction-based or classification-based methods. We adopt STN to align the images and Siamese network to implement the comparison. The SOTA results achieved on MVTec and MPDD show the effectiveness of our method.

\section{Experiments}

\subsection{Experiments with a Large k.} 
To decrease the training burden, RegAD is designed to be adaptable to unseen categories without parameter fine-tuning. Without fine-tuning on the few-shot support examples, simply increasing the shot number, the performance gain saturates very soon. We further experiment with k=64 and k=128. As shown in Table~\ref{tab:largeK}, when k increases from 64 to 128, a limited performance gain is observed. But the results are still competitive, though with a shallow backbone, compared to those of the AD methods trained by full data.

\begin{table}[h]
    \centering
    \scriptsize
    \caption{Comparison with AD method trained by full data.}
    \begin{tabular}{C{1.7cm}|C{1.5cm}C{1.5cm}C{1.5cm}C{2.1cm}C{2.1cm}}
        \toprule
         &  \multicolumn{3}{c}{RegAD} & PatchCore~\cite{patchcore} & CflowAD~\cite{cflow}\\
         \hline
       k  &  32 & 64 & 128 & full data & full data\\
       \hline
       MVTec & 94.6\% & 95.5\% & 95.9\% & 99.1\% & 98.3\%\\
       MPDD & 76.8\% & 82.3\% & 83.2\% & 82.1\% & 86.1\%\\
       \bottomrule
    \end{tabular}
    \label{tab:largeK}
\end{table}

\begin{table}[t]
\centering
\caption{Ablation studies of different versions of support set augmentations on the MVTec and MPDD datasets with $k=2$. Besides the full version of RegAD, We also provide the individual training version to reduce the influence of data augmentations on multiple categories. G, F, T, R means graying, flipping, translation, and rotation, respectively. Results are listed as the macro-average AUC in \% over all categories in each dataset of 10 runs. The best-performing method is in bold.}
\label{tal:abl_aug}
\scriptsize
\setlength{\tabcolsep}{0.6pt}{
\begin{tabular}{C{0.7cm}C{0.7cm}C{0.7cm}C{0.7cm}C{1.1cm}C{1.1cm}C{1.1cm}C{1.1cm}C{1.1cm}C{1.1cm}C{1.1cm}C{1.1cm}}
\toprule
\multicolumn{4}{c}{\multirow{2}{*}{Augmentations}} & \multicolumn{4}{c}{Individual Training} & \multicolumn{4}{c}{Aggregated Training}\\
\cmidrule(lr){5-8} \cmidrule(lr){9-12}
&&&& \multicolumn{2}{c}{MVTec} & \multicolumn{2}{c}{MPDD} & \multicolumn{2}{c}{MVTec} & \multicolumn{2}{c}{MPDD}\\
G & F & T & R & image & pixel & image & pixel & image & pixel & image & pixel\\
\cmidrule(lr){1-1} \cmidrule(lr){2-2} \cmidrule(lr){3-3} \cmidrule(lr){4-4} \cmidrule(lr){5-5} \cmidrule(lr){6-6} \cmidrule(lr){7-7} \cmidrule(lr){8-8} \cmidrule(lr){9-9} \cmidrule(lr){10-10} \cmidrule(lr){11-11} \cmidrule(lr){12-12}
 & & & & 74.7 & 88.6 & 49.6 & 89.5 & 79.1 & 90.5 & 57.6 & 91.0\\
\checkmark& & & & 74.9 & 88.6 & 49.4 & 89.4 & 79.5 & 90.7 & 58.0 & 91.3\\
 &\checkmark& & & 75.5 & 90.0 & 50.1 & 90.2 & 79.8 & 92.6 & 58.6 & 92.6\\
 & &\checkmark& & 77.4 & 90.9 & 49.8 & 91.5 & 81.3 & 92.4 & 58.3 & 90.9\\
 & & &\checkmark& 79.6 & 92.7 & 50.0 & 91.7 & 82.2 & 93.6 & 59.8 & 90.8\\
 \cmidrule(lr){1-1} \cmidrule(lr){2-2} \cmidrule(lr){3-3} \cmidrule(lr){4-4} \cmidrule(lr){5-5} \cmidrule(lr){6-6} \cmidrule(lr){7-7} \cmidrule(lr){8-8} \cmidrule(lr){9-9} \cmidrule(lr){10-10} \cmidrule(lr){11-11} \cmidrule(lr){12-12}
\checkmark&\checkmark& & & 75.6 & 89.9 & 50.0 & 90.1 & 80.5 & 92.7 & 59.7 & 91.7\\
\checkmark& &\checkmark& & 77.5 & 90.9 & 49.8 & 91.4 & 81.0 & 92.4 & 58.5 & 92.6\\
\checkmark& & &\checkmark& 79.7 & 92.6 & 50.0 & 91.6 & 83.8 & 93.7 & 60.9 & 92.6\\
 &\checkmark&\checkmark& & 77.7 & 91.5 & 50.1 & 91.7 & 81.6 & 93.2 & 58.2 & 92.3\\
 &\checkmark& &\checkmark& 79.7 & 92.9 & 51.3 & 91.8 & 82.3 & 94.0 & 59.7 & 92.9\\
 & &\checkmark&\checkmark& 81.3 & 93.1 & 49.9 & 92.2 & 84.2 & 94.6 & 60.6 & 91.7\\
 \cmidrule(lr){1-1} \cmidrule(lr){2-2} \cmidrule(lr){3-3} \cmidrule(lr){4-4} \cmidrule(lr){5-5} \cmidrule(lr){6-6} \cmidrule(lr){7-7} \cmidrule(lr){8-8} \cmidrule(lr){9-9} \cmidrule(lr){10-10} \cmidrule(lr){11-11} \cmidrule(lr){12-12}
\checkmark&\checkmark&\checkmark& & 77.8 & 91.5 & 50.2 & 91.6 & 81.7 & 93.5 & 59.9 & 92.6\\
\checkmark&\checkmark& &\checkmark& 79.8 & 92.9 & 51.2 & 91.7 & 83.9 & 94.2 & 63.0 & 93.0\\
\checkmark& &\checkmark&\checkmark& 81.4 & 93.1 & 49.9 & 92.3 & 84.9 & \textbf{94.7} & 61.2 & 92.8\\
&\checkmark&\checkmark&\checkmark& \textbf{81.5} & \textbf{93.3} & 50.7 & \textbf{92.4} & 85.4 & 94.6 & 61.2 & \textbf{93.2}\\
\cmidrule(lr){1-1} \cmidrule(lr){2-2} \cmidrule(lr){3-3} \cmidrule(lr){4-4} \cmidrule(lr){5-5} \cmidrule(lr){6-6} \cmidrule(lr){7-7} \cmidrule(lr){8-8} \cmidrule(lr){9-9} \cmidrule(lr){10-10} \cmidrule(lr){11-11} \cmidrule(lr){12-12}
\checkmark&\checkmark&\checkmark&\checkmark& \textbf{81.5} & \textbf{93.3} & \textbf{50.8} & \textbf{92.4} & \textbf{85.7} & 94.6 & \textbf{63.4} & \textbf{93.2}\\
\bottomrule
\end{tabular}}
\end{table}

\subsection{Ablation Studies on Support Set Augmentations.}
The proposed support set augmentations are shown to be essential for both detection and localization. Table~\ref{tal:abl_aug} further presents the ablation studies of comparing different augmentation methods for support images with $k=2$. The experimental results have validated the effectiveness of all the proposed augmentation methods. In particular, rotation and translation are shown to perform better on MVTec, while flipping and rotation seem to perform better on MPDD.

\subsection{Comparisons with Metaformer~\cite{metaformer}.}
Although using different training settings, according to the reported results in~\cite{metaformer}, Metaformer achieves about 88\% AUC for MVTec when k=8, while RegAD achieves 91.2\% AUC, an $\approx$3\% improvement, with the same test set and evaluation protocol. Metaformer achieves worse results despite three unfair advantages: (i) an additional large-scale dataset, MSRA10K, is used during its entire meta-training procedure (beyond parameter pre-training), together with additional pixel-level annotations; (ii) it performs additional fine-tuning on each novel category; (iii) it is trained with a deep transformer architecture for more epochs (100 vs. 50), with a larger batch size (64 vs. 32).

\end{document}